\documentclass[pdflatex,sn-basic,iicol]{sn-jnl}% Default with double column layout

\usepackage{amsmath}
\usepackage{graphicx}
\usepackage{natbib}
\usepackage{url}

\usepackage{graphicx,psfrag,epsf}
\usepackage{enumerate}

\usepackage{csquotes}
\usepackage{xspace}
\usepackage{subcaption}
\usepackage{enumitem}

\usepackage{amssymb,amsfonts}

\usepackage{algorithm}
\usepackage{float}
\usepackage{url}

\algblockdefx[AtSite]{AtSite}{Send}%
    [1][Site]{\textbf{At #1 do}}%
    [1][Send]{\textbf{Send} #1}

\usepackage{mathtools}
\usepackage{bm} %basic-ml, ml-gp
\usepackage{siunitx} %basic-ml
\usepackage{dsfont} %basic-math
\usepackage{xspace} %ml-mbo

\ifdefined\N                                                                
\renewcommand{\N}{\mathds{N}} % N, naturals
\else \newcommand{\N}{\mathds{N}} \fi 
\renewcommand{\R}{\mathds{R}} % R, reals
\ifdefined\C 
  \renewcommand{\C}{\mathds{C}} % C, complex
\else \newcommand{\C}{\mathds{C}} \fi
\newcommand{\argmin}{\operatorname{arg\,min}} % argmin
\newcommand{\diag}{\operatorname{diag}} % diag, diagonal
\newcommand{\Xspace}{\mathcal{X}} % X, input space
\newcommand{\Yspace}{\mathcal{Y}} % Y, output space
\newcommand{\Pxy}{\mathbb{P}_{xy}} % P_xy
\newcommand{\xv}{\bm{x}} % vector x (bold)
\newcommand{\yv}{\bm{y}} % vector y (bold)
\newcommand{\xy}{(\xv, y)} % observation (x, y)
\newcommand{\D}{\mathcal{D}} % D, data
\newcommand{\xyi}[1][i]{\left(\xv^{(#1)}, y^{(#1)}\right)} % (x^i, y^i), i-th observation
\newcommand{\Dset}{\left( \xyi[1], \ldots, \xyi[n]\right)} % {(x1,y1)), ..., (xn,yn)}, data
\renewcommand{\xi}[1][i]{\xv^{(#1)}} % x^i, i-th observed value of x
\newcommand{\xj}{\xv_j} % x_j, j-th feature
\newcommand{\fh}{\hat{f}} % f hat, estimated prediction function
\newcommand{\thetabh}{\bm{\hat\theta}} % theta vector hat
\newcommand{\riske}{\mathcal{R}_{\text{emp}}} % R_emp, empirical risk w/o factor 1 / n
\newcommand{\fmh}[1][m]{\hat{f}^{[#1]}} % prediction in iteration m
\newcommand{\fmdh}[1][m]{\hat{f}^{[#1-1]}} % prediction m-1
\newcommand{\rmm}[1][m]{\tilde{r}^{[#1]}} % pseudo residuals
\newcommand{\rmi}[1][m]{\tilde{r}^{[#1](i)}} % pseudo residuals
\renewcommand{\Dset}{\left\{ \xyi[1], \ldots, \xyi[n]\right\}} 
\newcommand{\xyik}[1][i]{\left(\xv^{(#1)}_\iSite, y^{(#1)}_\iSite\right)}
\newcommand{\Dsetk}[1][\iSite]{\left\{ \xyik[1], \ldots, \xyik[n_#1]\right\}}

\newcommand*{\tran}{{\mkern-1.5mu\mathsf{T}}}

\renewcommand{\th}[1][th]{\vphantom{x}^{\text{#1}}}

\newcommand{\tb}{\bm{\theta}}

\newcommand{\tbih}[1]{\hat{\tb}^{[#1]}}

\newcommand{\tbh}{\hat{\tb}}
\newcommand{\tbmh}{\tbh^{[m]}}

\newcommand{\bj}{b_l}
\newcommand{\bjm}{b_{l^{[m]}}}

\renewcommand{\rmm}[1][m]{\tilde{\bm{r}}^{[#1]}}

\newcommand{\hp}{\bm{\alpha}}

\newcommand{\lsite}[1][l]{#1_\times}
\newcommand{\blsite}[1][l]{b_{\lsite[#1]}}

\newcommand{\Zmat}{\bm{Z}}
\newcommand{\design}{\Zmat}

\newcommand{\xij}[1][i]{x_j^{(#1)}}
\renewcommand{\xj}[1][j]{\xv_#1}

\newcommand{\kron}{\times}

\newcommand{\penMat}{\bm{K}}
\newcommand{\idMat}{\bm{I}}
\newcommand{\nSites}{S}
\newcommand{\iSite}{s}
\newcommand{\sse}{\operatorname{SSE}}

\makeatletter
\renewcommand\operator@font{\sf}
\makeatother

\newcommand{\doH}{\texttt{[H]}\xspace}
\newcommand{\doS}{\texttt{[S]}\xspace}
\newcommand{\algospace}{\hspace{\algorithmicindent}}

\jyear{2023}%

\theoremstyle{thmstyleone}%
\theoremstyle{thmstyletwo}%

\theoremstyle{thmstylethree}%

\begin{document}

\title[ ]{Privacy-Preserving and Lossless Distributed Estimation of High-Dimensional Generalized Additive Mixed Models}

%% ==================================================== %%
%% AUTHORS - Stat & Computing                           %%
%% ==================================================== %%

\author*[1,2]{\fnm{Schalk} \sur{Daniel}}\email{daniel.schalk@stat.uni-muenchen.de}

\author[1,2]{\fnm{Bischl} \sur{Bernd}}\email{bernd.bischl@stat.uni-muenchen.de}

\author[1,2,3]{\fnm{R{\"u}gamer} \sur{David}}\email{david.ruegamer@stat.uni-muenchen.de}

\affil*[1]{\orgdiv{Department of Statistics}, \orgname{LMU Munich}, \orgaddress{\city{Munich}, \country{Germany}}}

\affil[2]{\orgdiv{Munich Center for Machine Learning (MCML)}}

\affil[3]{\orgdiv{Department of Statistics}, \orgname{TU Dortmund}, \orgaddress{\city{Dortmung}, \country{Germany}}}

\abstract{Various privacy-preserving frameworks that respect the individual's privacy in the analysis of data have been developed in recent years. However, available model classes such as simple statistics or generalized linear models lack the flexibility required for a good approximation of the underlying data-generating process in practice. In this paper, we propose an algorithm for a distributed, privacy-preserving, and lossless estimation of generalized additive mixed models (GAMM) using component-wise gradient boosting (CWB). Making use of CWB allows us to reframe the GAMM estimation as a distributed fitting of base learners using the $L_2$-loss. In order to account for the heterogeneity of different data location sites, we propose a distributed version of a row-wise tensor product that allows the computation of site-specific (smooth) effects. Our adaption of CWB preserves all the important properties of the original algorithm, such as an unbiased feature selection and the feasibility to fit models in high-dimensional feature spaces, and yields equivalent model estimates as CWB on pooled data. Next to a derivation of the equivalence of both algorithms, we also showcase the efficacy of our algorithm on a distributed heart disease data set and compare it with state-of-the-art methods.}


\maketitle

\section{Introduction}

More than ever, data is collected to record the ubiquitous information in our everyday life. However, on many occasions, the physical location of data points is not confined to one place (one global site) but distributed over different locations (sites). This is the case for, e.g., patient records that are gathered at different hospitals but usually not shared between hospitals or other facilities due to the sensitive information they contain. This makes data analysis challenging, particularly if methods require or notably benefit from incorporating all available (but distributed) information. For example, personal patient information is typically distributed over several hospitals, while sharing or merging different data sets in a central location is prohibited. To overcome this limitation, different approaches have been developed to directly operate at different sites and unite information without having to share sensitive parts of the data to allow privacy-preserving data analysis.

\paragraph{Distributed Data} Distributed data can be partitioned vertically or horizontally across different sites. Horizontally partitioned data means that observations are spread across different sites with access to all existing features of the available data point, while for vertically partitioned data, different sites have access to all observations but different features (covariates) for each of these observations. In this work, we focus on horizontally partitioned data. Existing approaches for horizontally partitioned data vary from fitting regression models such as generalized linear models~\citep[GLMs;][]{wu2012g, lu2015webdisco, jones2013combined, chen2018privacy}, to conducting distributed evaluations~\citep{boyd2015differential, unal2021ppaurora, schalk2022distributed}, to fitting artificial neural networks~\citep{mcmahan2017communication}. Furthermore, various software frameworks are available to run a comprehensive analysis of distributed data. One example is the collection of \texttt{R}~\citep{rcore} packages \texttt{DataSHIELD}~\citep{gaye2014datashield}, which enables data management and descriptive data analysis as well as securely fitting of simple statistical models in a distributed setup without leaking information from one site to the others.

\paragraph{Interpretability and Data Heterogeneity} In many research areas that involve critical decision-making, especially in medicine, methods should not only excel in predictive performance but also be interpretable. Models should provide information about the decision-making process, the feature effects, and the feature importance as well as intrinsically select important features. Generalized additive models \citep[GAMs; see, e.g.,][]{Wood.2017.book} are one of the most flexible approaches in this respect, providing an interpretable yet complex models that also allow for non-linearity in the data. 

As longitudinal studies are often the most practical way to gather information in many research fields, methods should also be able to account for subject-specific effects and account for the correlation of repeated measurements. Furthermore, when analyzing data originating from different sites, the assumption of having identically distributed observations across all sites often does not hold. In this case, a reasonable assumption for the data-generating process is a site-specific deviation from the general population mean. Adjusting models to this situation is called interoperability~\citep{litwin1990interoperability}, while ignoring it may lead to biased or wrong predictions. 

\subsection{Related Literature} \label{sec:rellit}

Various approaches for distributed and privacy-preserving analysis have been proposed in recent years. In the context of statistical models, \citet{karr2005secure} describe how to calculate a linear model (LM) in a distributed and privacy-preserving fashion by sharing data summaries. \citet{jones2013combined} propose a similar approach for GLMs by communicating the Fisher information and score vector to conduct a distributed Fisher scoring algorithm. The site information is then globally aggregated to estimate the model parameters. Other privacy-preserving techniques include ridge regression~\citep{chen2018privacy}, logistic regression, and neural networks~\citep{mohassel2017secureml}. 

In machine learning, methods such as the naive Bayes classifier, trees, support vector machines, and random forests~\citep{li2020privacy} exist with specific encryption techniques~\citep[e.g., the Paillier cryptosystem;][]{paillier1999public} to conduct model updates. In these setups, a trusted third party is usually required. However, this is often unrealistic and difficult to implement, especially in a medical or clinical setup. Furthermore, as encryption is an expensive operation, its application is infeasible for complex algorithms that require many encryption calls~\citep{naehrig2011can}. Existing privacy-preserving boosting techniques often focus on the AdaBoost algorithm by using aggregation techniques of the base classifier~\citep{lazarevic2001distributed,gambs2007privacy}. A different approach to boosting decision trees in a federated learning setup was introduced by~\citep{li2020practical} using a locality-sensitive hashing to obtain similarities between data sets without sharing private information. These algorithms focus on aggregating tree-based base components, making them difficult to interpret, and come with no inferential guarantees. 

In order to account for repeated measurements, \citet{luo2022dlmm} propose a privacy-preserving and lossless way to fit linear mixed models (LMMs) to correct for heterogeneous site-specific random effects. Their concept of only sharing aggregated values is similar to our approach, but is limited in the complexity of the model and only allows normally distributed outcomes. Other methods to estimate LMMs in a secure and distributed fashion are~\citet{zhu2020distlmm},~\citet{anjum2022privacy}, or~\citet{yan2022fed}. 

Besides privacy-preserving and distributed approaches, integrative analysis is another technique based on pooling the data sets into one and analyzing this pooled data set while considering challenges such as heterogeneity or the curse of dimensionality~\citep{curran2009integrative, bazeley2012integrative, mirza2019machine}. While advanced from a technical perspective by, e.g., outsourcing computational demanding tasks such as the analysis of multi-omics data to cloud services \citep{elab007}, the existing statistical cloud-based methods only deal with basic statistics. The challenges of integrative analysis are similar to the ones tackled in this work, our approach, however, does not allow merging the data sets in order to preserve privacy. 

\subsection{Our Contribution}

This work presents a method to fit generalized additive mixed models (GAMMs) in a privacy-preserving and lossless manner\footnote{In this article, we define a distributed fitting procedure as lossless if the model parameters of the algorithm are the same as the ones computed on the pooled data.} to horizontally distributed data. This not only allows the incorporation of site-specific random effects and accounts for repeated measurements in LMMs, but also facilitates the estimation of mixed models with responses following any distribution from the exponential family and provides the possibility to estimate complex non-linear relationships between covariates and the response. To the best of our knowledge, we are the first to provide an algorithm to fit the class of GAMMs in a privacy-preserving and lossless fashion on distributed data.

Our approach is based on component-wise gradient boosting~\citep[CWB;][]{buhlmann2003boosting}. CWB can be used to estimate additive models, account for repeated measurements, compute feature importance, and conduct feature selection. Furthermore, CWB is suited for high-dimensional data situations $(n\ll p)$. CWB is therefore often used in practice for, e.g., predicting the development of oral cancer~\citep{saintigny2011gene}, classifying individuals with and without patellofemoral pain syndrome~\citep{liew2020classifyingpatella}, or detecting synchronization in bioelectrical signals~\citep{rugamer2016boosting}. However, there have so far not been any attempts to allow for a distributed, privacy-preserving, and lossless computation of the CWB algorithm. In this paper, we propose a distributed version of CWB that yields the identical model produced by the original algorithm on pooled data and that accounts for site heterogeneity by including interactions between features and a site variable. This is achieved by adjusting the fitting process using 1) a distributed estimation procedure, 2) a distributed version of row-wise tensor product base learners, and 3) an adaption of the algorithm to conduct feature selection in the distributed setup. We implement our method in \texttt{R} using the \texttt{DataSHIELD} framework and demonstrate its application in an exemplary medical data analysis. Our distributed version of the original CWB algorithm does not have any additional HPs and uses optimization strategies from previous research results to define meaningful values for all hyperparameters, effectively yielding a tuning-free method.

The remainder of this paper is structured as follows: First, we introduce the basic notation, terminology, and setup of GAMMs in Section~\ref{sec:background}. We then describe the original CWB algorithm in Section~\ref{sec:methods} and its link to GAMMs. In Section~\ref{sec:distr-cwb}, we present the distributed setup and our novel extension of the CWB algorithm. Finally, Section~\ref{sec:application} demonstrates both how our distributed CWB algorithm can be used in practice and how to interpret the obtained results. 

\paragraph{Implementation} We implement our approach as an \texttt{R} package using the DataSHIELD framework and make it available on GitHub\footnote{\url{github.com/schalkdaniel/dsCWB}}. The code for the analysis can also be found in the repository\footnote{\url{github.com/schalkdaniel/dsCWB/blob/main/usecase/analyse.R}}. 

\section{Background}\label{sec:background}

\subsection{Notation and Terminology}

Our proposed approach uses the CWB algorithm as fitting engine. Since this method was initially developed in machine learning, we introduce here both the statistical notation used for GAMMs as well as the respective machine learning terminology and explain how to relate the two concepts. 

We assume a $p$-dimensional covariate or feature space $\Xspace = (\Xspace_1 \times \hdots \times \Xspace_p) \subseteq \mathbb{R}^p$ and response or outcome values from a target space $\Yspace$. The goal of boosting is to find the unknown relationship $f$ between $\Xspace$ and $\Yspace$. In turn, GAMMs (as presented in Section~\ref{sec:GAMM}) model the conditional distribution of an outcome variable $Y$ with realizations $y\in\Yspace$, given features $\xv = (x_1, \dots, x_p) \in \Xspace$. Given a data set $\D = \Dset$ with $n$ observations drawn (conditionally) independently from an unknown probability distribution $\Pxy$ on the joint space $\Xspace \times \Yspace$, we aim to estimate this functional relationship in CWB with $\fh$. The goodness-of-fit of a given model $\fh$ is assessed by calculating the empirical risk $\riske(\fh) = n^{-1}\sum_{\xy \in \mathcal{D}}L(y, \fh(\xv))$ based on a loss function $L: \Yspace \times \R \to \R$ and the data set $\D$. Minimizing $\riske$ using this loss function is equivalent to estimating $f$ using maximum likelihood by defining $L(y,f(\bm{x})) = -\ell(y,h(f(\bm{x})))$ with log-likelihood $\ell$, response function $h$ and minimizing the sum of log-likelihood contributions.

In the following, we also require the vector $\xj = (\xij[1], \ldots, \xij[n])^\tran \in \Xspace_j$, which refers to the $j\th$ feature. Furthermore, let $\xv = (x_1, \dots, x_p)$ and $y$ denote arbitrary members of $\Xspace$ and $\Yspace$, respectively. A special role is further given to a subset $\bm{u} = (u_1,\ldots,u_q)^\top$, $q\leq p$, of features $\xv$, which will be used to model the heterogeneity in the data. 

\subsection{Generalized Additive Mixed Models} \label{sec:GAMM}

A very flexible class of regression models to model the relationship between covariates and the response are GAMMs~\citep[see, e.g.,][]{Wood.2017.book}. In GAMMs, the response $Y^{(i)}$ for observation $i=1,\ldots,n_s$ of measurement unit (or site) $s$ is assumed to follow some exponential family distribution such as the Poisson, binomial, or normal distributions \citep[see, e.g.,][]{mccullagh2019generalized}, conditional on features $\bm{x}^{(i)}$ and the realization of some random effects. The expectation $\mu := \mathbb{E}(Y^{(i)} \vert \bm{x}^{(i)}, \bm{u}^{(i)})$ of the response $Y^{(i)}$ for observations $i=1,\ldots,n_s$ of measurement unit (or site) $s$ in GAMMs is given by
\begin{align} 
    h^{-1}&(\mu^{(i)}) = f^{(i)} \notag \\
    &= \sum_{j\in\mathcal{J}_{1}} x^{(i)}_j \beta_j + \sum_{j\in\mathcal{J}_{2}} u^{(i)}_j \gamma_{j,s} + \sum_{j\in\mathcal{J}_{3}} \phi_j(x^{(i)}_j). \label{eq:GAMM}
\end{align}
In \eqref{eq:GAMM}, $h$ is a smooth monotonic response function, $f$ corresponds to the additive predictor, $\gamma_{j,s} \sim \mathcal{N}(0, \psi)$ are random effects accounting for heterogeneity in the data, and $\phi_j$ are non-linear effects of pre-specified covariates. The different index sets $\mathcal{J}_1, \mathcal{J}_2, \mathcal{J}_3 \subseteq \{1,\ldots,p\} \cup \emptyset$ indicate which features are modeled as fixed effects, random effects, or non-linear (smooth) effects, respectively. The modeler usually defines these sets. However, as we will also explain later, the use of CWB as a fitting engine allows for automatic feature selection and therefore does not require explicitly defining these sets. In GAMMs, smooth effects are usually represented by (spline) basis functions, i.e., $\phi_j(x_j) \approx (B_{j,1}(x_j), \ldots, B_{j,d_j}(x_j))^\top \bm{\theta}_j$, where $\bm{\theta}_j \in \mathbb{R}^{d_j}$ are the basis coefficients corresponding to each basis function $B_{j,d_j}$. The coefficients are typically constrained in their flexibility by adding a quadratic (difference) penalty for (neighboring) coefficients to the objective function to enforce smoothness. GAMMs, as in \eqref{eq:GAMM}, are not limited to univariate smooth effects $\phi_j$, but allow for higher-dimensional non-linear effects $\phi(x_{j_1}, x_{j_2}, \ldots, x_{j_k})$. The most common higher-dimensional smooth interaction effects are bivariate effects ($k=2$) and can be represented using a bivariate or a tensor product spline basis (see Section~\ref{subsubsec:base-learner} for more details). Although higher-order splines with $k>2$ are possible, models are often restricted to bivariate interactions for the sake of interpretability and computational feasibility. In Section~\ref{sec:distr-cwb}, we will further introduce varying coefficient terms $\phi_{j,\iSite}(x_j)$ in the model \eqref{eq:GAMM}, i.e., smooth effects $f$ varying with a second variable $\iSite$. Analogous to random slopes, $\iSite$ can also be the index set defining observation units of random effects $\mathcal{J}_2$. Using an appropriate distribution assumption for the basis coefficients $\bm{\theta}_j$, these varying coefficients can then be considered as random smooth effects.

\subsection{Component-Wise Boosting}~\label{sec:methods}

Component-wise (gradient) boosting~\citep[CWB;][]{buhlmann2003boosting,buhlmann2007boosting} is a an iterative algorithm that performs block-coordinate descent steps with blocks (or base learners) corresponding to the additive terms in \eqref{eq:GAMM}. With a suitable choice of base learners and objective function, CWB allows efficient optimization of GAMMs, even in high-dimensional settings with $p \gg n$. We will first introduce the concept of base learners that embed additive terms of the GAMM into boosting and subsequently describe the actual fitting routine of CWB. Lastly, we will describe the properties of the algorithm and explain its connection to model \eqref{eq:GAMM}. 

\subsubsection{Base Learners}\label{subsubsec:base-learner}

In CWB, the $l\th$ base learner $b_l: \Xspace \to \R$ is used to model the contribution of one or multiple features in the model. In this work, we investigate parametrized base learners $b_l(\xv, \tb_l)$ with parameters $\tb_l\in\R^{d_l}$. For simplicity, we will use $\bm{\theta}$ as a wildcard for the coefficients of either fixed effects, random effects, or spline bases in the following. We assume that each base learner can be represented by a generic basis representation $g_l : \Xspace \to \R^{d_l},\ \xv \mapsto g_l(\xv) = (g_{l,1}(\xv), \dots, g_{l, d_l}(\xv))^\tran$ and is linear in the parameters, i.e., $b_l(\xv, \tb_l) = g_l(\xv)^\tran \tb_l$. For $n$ observations, we define the design matrix of a base learner $b_l$ as $\design_l := (g_l(\xi[1]), \ldots, g_l(\xi[n]))^\tran\in\R^{n\times d_l}$. Note that base learners are typically not defined on the whole feature space but on a subset $\Xspace_l\subseteq\Xspace$. For example, a common choice for CWB is to define one base learner for every feature $x_l\in\mathcal{X}_l$ to model the univariate contributions of that feature. 

A base learner $b_l(\xv, \tb_l)$ can depend on HPs $\hp_l$ that are set prior to the fitting process. For example, choosing a base learner using a P-spline \citep{eilers1996flexible} representation requires setting the degree of the basis functions, the order of the difference penalty term, and a parameter $\lambda_l$ determining the smoothness of the spline. In order to represent GAMMs in CWB, the following four base learner types are used.

\paragraph{(Regularized) linear base learners} A linear base learner is used to include linear effects of a features $x_{j_1}, \dots, x_{j_{d_l}}$ into the model. The basis transformation is given by $g_l(\xv) = (g_{l,1}(\xv), \dots, g_{l,d_l + 1}(\xv))^\tran = (1, x_{j_1}, \dots, x_{j_{d_l}})^\tran$. Linear base learners can be regularized by incorporating a ridge penalization~\citep{hoerl1970ridge} with tunable penalty parameter $\lambda_l$ as an HP $\hp_l$. Fitting a ridge penalized linear base learner to a response vector $\yv \in \mathbb{R}^n$ results in the penalized least squares estimator $\tbh_l = (\Zmat_l^\tran \Zmat_l + \penMat_l)^{-1}\Zmat_l^\tran \yv$ with penalty matrix $\penMat_l = \lambda_l\idMat_{d_l + 1}$, where $\idMat_d$ denotes the $d$-dimensional identity matrix. Often, an unregularized linear base learner is also included to model the contribution of one feature $\xj$ as a linear base learner without penalization. The basis transformation is then given by $g_l(\xv) = (1, x_j)^\tran$ and $\lambda_l = 0$.
        
\paragraph{Spline base learners} These base learners model smooth effects using univariate splines. A common choice is penalized B-splines~\citep[P-Splines;][]{eilers1996flexible}, where the feature $x_j$ is transformed using a B-spline basis transformation $g_l(\xv) = (B_{l,1}(x_j), \dots, B_{l,d_l}(x_j))^\tran$ with $d_l$ basis functions $g_{l,m} = B_{l,m},\ m = 1, \dots, d_l$. In this case, the choice of the spline order $B$, the number of basis functions $d_l$, the penalization term $\lambda_l$, and the order $v$ of the difference penalty (represented by a matrix $\bm{D}_l\in\R^{d_{l-v}\times d_l}$) are considered HPs $\hp_l$ of the base learner. The base learner's parameter estimator in general is given by the penalized least squares solution $\tbh_l = (\Zmat_l^\tran \Zmat_l + \penMat_l)^{-1}\Zmat_l^\tran \yv$, with penalization matrix $\penMat_l = \lambda_l\bm{D}_l^\top \bm{D}_l$ in the case of P-splines.
        
\paragraph{Categorical and random effect base learners} Categorical features $x_j\in\{1, \dots, G\}$ with $G\in\N, G\geq 2$ classes are handled by a binary encoding $g_l(\xv) = (\mathds{1}_{\{1\}}(x_j), \dots, \mathds{1}_{\{G\}}(x_j))^\tran$ with the indicator function $\mathds{1}_A(x) = 1$ if $x\in A$ and $\mathds{1}_A(x) = 0$ if $x\notin A$. A possible alternative encoding is the dummy encoding with $\breve g_l(\xv) = (1, \mathds{1}_{\{1\}}(x_j), \dots, \mathds{1}_{\{G-1\}}(x_j))^\tran$ with reference group $G$. Similar to linear and spline base learners, it is possible incorporate a ridge penalization with HP $\hp_l=\lambda_l$. This results in the base learner's penalized least squared estimator $\tbh_l = (\Zmat_l^\tran \Zmat_l + \penMat_l)^{-1}\Zmat_l^\tran \yv$ with penalization matrix $\penMat_l = \lambda_l\idMat_{G}$. Due to the mathematical equivalence of ridge penalized linear effects and random effects with normal prior~\citep[see, e.g.,][]{brumback1999variable}, this base learner can further be used to estimate random effect predictions $\hat{\gamma}_j$ when using categorical features $u_j$ and thereby account for heterogeneity in the data.
        
\paragraph{Row-wise tensor product base learners} This type of base learner is used to model a pairwise interaction between two features $x_j$ and $x_k$. Given two base learners $b_j$ and $b_k$ with basis representations $g_j(\xv) = (g_{j,1}(x_j), \dots, g_{j,d_j}(x_j))^\tran$ and $g_k(\xv) = (g_{k,1}(x_k), \dots, g_{k,d_k}(x_k))^\tran$, the basis representation of the row-wise tensor product base learner $b_l = b_j \times b_k$ is defined as $g_l(\xv) = (g_j(\xv)^\tran \otimes g_k(\xv)^\tran)^\tran = (g_{j,1}(x_j) g_k(\xv)^\tran, \dots, g_{j,d_j}(x_j) g_k(\xv)^\tran)^\tran\in\R^{d_l}$ with $d_l = d_j d_k$. The HPs $\hp_l = \{\hp_j, \hp_k\}$ of a row-wise tensor product base learner are induced by the HPs $\hp_j$ and $\hp_k$ of the respective individual base learners. Analogously to other base learners, the penalized least squared estimator in this case is $\tbh_l = (\Zmat_l^\tran \Zmat_l + \penMat_l)^{-1}\Zmat_l^\tran \yv$ with penalization matrix $\penMat_l = \tau_j \penMat_j \otimes \idMat_{d_k} + \idMat_{d_j} \otimes \tau_k \penMat_k \in\R^{d_l \times d_l}$. This Kronecker sum penalty, in particular, allows for anisotropic smoothing with penalties $\tau_j$ and $\tau_k$ when using two spline bases for $g_j$ and $g_k$, and varying coefficients or random splines when combining a (penalized) categorical base learner and a spline base learner.

\subsubsection{Fitting Algorithm}\label{subsubsec:fitting-algo}

CWB first initializes an estimate $\fh$ of the additive predictor with a loss-optimal constant value $\fh^{[0]} = \argmin_{c\in\R}\riske(c)$. It then proceeds and estimates Eq.~\eqref{eq:GAMM} using an iterative steepest descent minimization in function space by fitting the previously defined base learners to the model's functional gradient $\nabla_f L(y,f)$ evaluated at the current model estimate $\hat{f}$. Let $\fmh$ denote the model estimation after $m\in\mathbb{N}$ iterations. In each step in CWB, the pseudo residuals $\rmi = -\nabla_f L(y^{(i)}, f(\xi))\rvert_{f = \fmdh}$ for $\  i \in \{1, \dots, n\}$ are first computed. CWB then selects the best-fitting base learner from a pre-defined pool of base-learners denoted by $\mathcal{B} = \{\bj\}_{l \in \{1, \ldots, \lvert\mathcal{B}\rvert\}}$ and adds the base learner's contribution to the previous model $\fmh$. The selected base learner is chosen based on its sum of squared errors (SSE) when regressing the pseudo residuals $\rmm = (\bm{r}^{[m](1)}, \dots, \bm{r}^{[m](n)})^\tran$ onto the base learner's features using the $L_2$-loss. Further details of CWB are given in Algorithm~\ref{algo:cwb} \citep[see, e.g.,][]{schalk2022accelerated}.

\paragraph{Controling HPs of CWB}

Good estimation performance can be achieved by selecting a sufficiently small learning rate, e.g., 0.01, as suggested in \citet{buhlmann2007boosting}, and adaptively selecting the number of boosting iterations via early stopping on a validation set. To enforce a fair selection of model terms and thus unbiased effect estimation, regularization parameters are set such that all base learners have the same degrees-of-freedom \citep{hofner2011framework}. As noted by~\citet{buhlmann2007boosting}, choosing smaller degrees-of-freedom induces more penalization (and thus, e.g., smoother estimated function for spline base learners), which yields a model with lower variance at the cost of a larger bias. This bias induces a shrinkage in the estimated coefficients towards zero but can be reduced by running the optimization process for additional iterations.

\begin{algorithm}[H]
%\footnotesize
\small
\caption{Vanilla CWB algorithm}\label{algo:cwb}
\vspace{0.15cm}
\hspace*{\algorithmicindent} \textbf{Input} Train data $\D$, learning rate $\nu$, number of\\
\hspace*{\algorithmicindent} \phantom{\textbf{Input} }boosting iterations $M$, loss function $L$,\\
\hspace*{\algorithmicindent} \phantom{\textbf{Input} }set of base learner $\mathcal{B}$\\
\hspace*{\algorithmicindent} \textbf{Output} Model $\hat{f}^{[M]}$ defined by fitted parameters\\
\hspace*{\algorithmicindent} \phantom{\textbf{Output} }$\tbih{1}, \ldots, \tbih{M}$\vspace{0.15cm}
\hrule

\begin{algorithmic}[1]
\Procedure{$\operatorname{CWB}$}{$\D,\nu,L,\mathcal{B}$}
    \State Initialize: $\fh^{[0]}(\xv) = \argmin_{c\in\R}\riske(c)$
    \For{$m \in \{1, \dots, M\}$}
        \State $\rmi = -\nabla_f L(y^{(i)}, f(\xi))\vert_{f = \fmdh}$,\ \ $\forall i \in \{1, \dots, n\}$\label{algo:cwb:line:pr}
        \For{$l \in \{1, \dots, \lvert\mathcal{B}\rvert\}$}
        \State $\tbmh_l = \left(\design_l^\tran \design_l + \penMat_l\right)^{-1} \design^\tran_l \rmm$\label{algo:cwb:line:fitbl} %\argmin_{\tb\in\R^{\bdj}} \sum_{i=1}^n\left(\rmi - \bj(\xi,\tb)\right)^2$
            \State $\sse_l = \sum_{i=1}^n(\rmi - \bj(\xi, \tbmh_l))^2$ \label{algo:cwb:line:sse}
        \EndFor
        \State $l^{[m]} = \argmin_{l\in\{1, \dots, \lvert\mathcal{B}\rvert\}} \sse_l$ \label{algo:cwb:line:blselection}
        \State $\fmh(\xv) = \fmdh(\xv) + \nu \bjm (\xv,\tbmh_{l^{[m]}})$
    \EndFor
    \State \textbf{return} $\fh = \fh^{[M]}$
\EndProcedure
\end{algorithmic}
\end{algorithm}

\subsubsection{Properties and Link to Generalized Additive Mixed Models}

The estimated coefficients $\hat{\bm{\theta}}$ resulting from running the CWB algorithm are known to converge to the maximum likelihood solution \citep[see, e.g.,][]{schmid2008} for $M\to \infty$ under certain conditions. This is due to the fact that CWB performs a coordinate gradient descent update of a model defined by its additive base learners that exactly represent the structure of an additive mixed model (when defining the base learners according to Section~\ref{subsubsec:base-learner}) and by the objective function that corresponds to the negative (penalized) log-likelihood. Two important properties of this algorithm are 1) its coordinate-wise update routine, and 2) the nature of model updates using the $L_2$-loss. Due to the first property, CWB can be used in settings with $p \gg n$, as only a single additive term is fitted onto the pseudo-residuals in every iteration. This not only reduces the computational complexity of the algorithm for an increasing number of additive predictors (linear instead of quadratic) but also allows variable selection when stopping the routine early (e.g., based on a validation data set), as not all the additive components might have been selected into the model. In particular, this allows users to specify the full GAMM model without manual specification of the type of feature effect (fixed or random, linear or non-linear) and then automatically sparsify this model by an objective and data-driven feature selection. The second property, allows fitting models of the class of \emph{generalized} linear/additive (mixed) models using only the $L_2$-loss instead of having to work with some iterative weighted least squares routine. In particular, this allows performing the proposed lossless distributed computations described in this paper, as we will discuss in Section~\ref{sec:distr-cwb}.

\subsection{Distributed Computing Setup and Privacy Protection} \label{subsec:dist-setup}

Before presenting our main results, we now introduce the distributed data setup we will work with throughout the remainder of this paper. The data set $\D$ is horizontally partitioned into $\nSites$ data sets $\D_\iSite = \Dsetk$, $\iSite = 1, \dots, \nSites$ with $n_\iSite$ observations. Each data set $\D_\iSite$ is located at a different site $\iSite$ and potentially follows a different data distributions $\mathbb{P}_{xy,\iSite}$. The union of all data sets yields the whole data set $\D = \cup_{\iSite=1}^\nSites \D_\iSite$ with mutually exclusive data sets $\D_\iSite \cap \D_l = \emptyset \,\,\forall l,\iSite \in \{1,\ldots,\nSites\}, l\neq \iSite$. The vector of realizations per site is denoted by $\yv_\iSite\in\Yspace^{n_\iSite}$.

In this distributed setup, multiple ways exist to communicate information without revealing individual information. More complex methods such as differential privacy~\citep{dwork2006differential}, homomorphic encryption \citep[e.g., the Paillier cryptosystem;][]{paillier1999public}, or k-anonymity~\citep{samarati1998protecting} allow sharing information without violating an individual's privacy. An alternative option is to only communicate aggregated statistics. This is one of the most common approaches and is also used by \texttt{DataSHIELD}~\citep{gaye2014datashield} for GLMs or by \citet{luo2022dlmm} for LMMs. \texttt{DataSHIELD}, for example, uses a privacy level that indicates how many individual values must be aggregated to allow the communication of aggregated values. For example, setting the privacy level to a value of 5 enables sharing of summary statistics such as sums, means, variances, etc. if these are computed on at least 5 elements (observations).

\paragraph{Host and Site Setup} Throughout this article, we assume the $1, \dots, \nSites$ sites or servers to have access to their respective data set $\D_\iSite$. Each server is allowed to communicate with a host server that is also the analyst's machine. In this setting, the analyst can potentially see intermediate data used when running the algorithms, and hence each message communicated from the servers to the host must not allow any reconstruction of the original data. The host server is responsible for aggregating intermediate results and communicating these results back to the servers.

\section{Distributed Component-Wise Boosting}\label{sec:distr-cwb}

We now present our distributed version of the CWB algorithm to fit privacy-preserving and lossless GAMMs.
%As a consequence, the distributed model $\fh_1$ and the model $\fh_2$ fitted on pooled data have the same empirical risk $\riske(\fh_1) = \riske(\fh_2)$. 
In the following, we first describe further specifications of our setup in Section~\ref{subsec:distr-setup}, elaborate on the changes made to the set of base learners in Section~\ref{subsec:distr-baselearner}, and then show how to adapt CWB's fitting routine in Section~\ref{subsec:dist-fitting-algo}. 

\subsection{Setup} \label{subsec:distr-setup}

In the following, we distinguish between site-specific and shared effects. As effects estimated across sites typically correspond to fixed effects and effects modeled for each site separately are usually represented using random effects, we use the terms as synonyms in the following, i.e., \emph{shared effects} and \emph{fixed effects} are treated interchangeably and the same holds for \emph{site-specific effects} and \emph{random effects}. We note that this is only for ease of presentation and our approach also allows for site-specific fixed effects and random shared effects. As the data is not only located at different sites but also potentially follows different data distributions $\mathbb{P}_{xy,\iSite}$ at each site $\iSite$, we extend Eq.~\eqref{eq:GAMM} to not only include random effects per site, but also site-specific smooth (random) effects $\phi_{j,\iSite}(x_j)$, $\iSite = 1, \dots, \nSites$ for all features $x_j$ with $j\in\mathcal{J}_3$. For every of these smooth effects $\phi_{j,\iSite}$ we assume an existing shared effect $f_{j,\text{shared}}$ that is equal for all sites. These assumptions -- particularly the choice of site-specific effects -- are made for demonstration purposes. In a real-world application, the model structure can be defined individually to match the given data situation. However, note again that CWB intrinsically performs variable selection, and there is thus no need to manually define the model structure in practice. In order to incorporate the site information into the model, we add a variable $x_0^{(i)} \in\{1, \dots, \nSites\}$ for the site to the data by setting $\tilde{\xv}^{(i)} = (x_0^{(i)}, \xv^{(i)})$. The site variable is a categorical feature with $\nSites$ classes. 

\subsection{Base Learners}\label{subsec:distr-baselearner}

For shared effects, we keep the original structure of CWB with base learners chosen from a set of possible learners $\mathcal{B}$. Section~\ref{subsubsec:shared-effects} explains how these shared effects are estimated in the distributed setup. 
We further define a regularized categorical base learner $b_0$ with basis transformation $g_0(x_0) = (\mathds{1}_{\{1\}}(x_0), \dots, \mathds{1}_{\{\nSites\}}(x_0))^\tran$ and design matrix $\design_0\in\R^{n\times \nSites}$. We use $b_0$ to extend $\mathcal{B}$ with a second set of base learners $\mathcal{B}_\times = \{b_0 \times b\ \vert\ b \in \mathcal{B}\}$ to model site-specific random effects. All base learners in $\mathcal{B}_\times$ are row-wise tensor product base learners $\blsite = b_0 \times b_l$ of the regularized categorical base learner $b_0$ dummy-encoding every site and all other existing base learners $b_l\in\mathcal{B}$. This allows for potential inclusion of random effects for every fixed effect in the model. More specifically, the $l\th$ site-specific effect given by the row-wise tensor product base learner $b_{\lsite}$ uses the basis transformation $g_{\lsite} = g_0 \otimes g_l$
\begin{align} 
    g_{\lsite}&(\tilde{\xv}) = g_0(x_0)^\tran \otimes g_l(\xv)^\tran \notag \\
    &= (\underbrace{\mathds{1}_{\{1\}}(x_0) g_l(\xv)^\tran}_{=g_{{\lsite}, 1}}, \dots, \underbrace{\mathds{1}_{\{\nSites\}}(x_0) g_l(\xv)^\tran}_{=g_{{\lsite}, \nSites}})^\tran, \label{eq:rowtensor-site}
\end{align}
where the basis transformation $g_l$ is equal for all $\nSites$ sites. After distributed computation (see Eq.~\eqref{eq:bl-fit-kronecker-pooled} in the next section), the estimated coefficients are $\thetabh_{\lsite} = (\thetabh_{{\lsite}, 1}^\tran, \dots, \thetabh_{{\lsite}, \nSites}^\tran)^\tran$ with $\thetabh_{{\lsite}, \iSite}\in\R^{d_l}$. The regularization of the row-wise Kronecker base learners not only controls their flexibility but also assures identifiable when additionally including a shared (fixed) effect for the same covariate. The penalty matrix $\penMat_{\lsite} = \lambda_0 \penMat_0 \otimes \idMat_{d_l} + \idMat_\nSites \otimes \lambda_{l_\times} \penMat_l\in\mathbb{R}^{\nSites d_l \times \nSites d_l}$ is given as Kronecker sum of the penalty matrix of the categorical site effect and the penalty matrices $\penMat_0$ and $\penMat_l$ with respective regularization strengths $\lambda_0, \lambda_{l_\times}$. As $\penMat_0 = \lambda_0\idMat_{\nSites}$ is a diagonal matrix, $\penMat_{\lsite}$ is a block matrix with entries $\lambda_0\idMat_{d_l} + \lambda_{l_\times} \penMat_l$ on the diagonal blocks. Moreover, as $g_0$ is a binary vector, we can also express the design matrix $\design_{\lsite}\in\mathbb{R}^{n\times \nSites d_l}$ as a block matrix, yielding
\begin{align} 
%    \design_{\lsite} = \underbrace{\left(
%    \begin{array}{ccc}
%         \design_{l,1} &        & \\
%                   & \ddots &  \\
%                   &        & \design_{l,\nSites}
%    \end{array}
%    \right)}_{\nSites\ \text{times}}, \
%    \penMat_{\lsite} = \underbrace{\left(
%    \begin{array}{ccc}
%         \lambda_0\idMat_{d_l} + \lambda_{l_\times}\penMat_l &        & \\
%                   & \ddots &  \\
%                   &        & \lambda_0\idMat_{d_l} + \lambda_{l_\times}\penMat_l
%    \end{array}
%    \right)}_{\nSites\ \text{times}},    
    \design_{\lsite} &= \diag(\design_{l,1}, \dots,  \design_{l,\nSites}),\ \penMat_{\lsite} \notag \\ &= \diag(\lambda_0\idMat_{d_l} + \lambda_{l_\times}\penMat_l, \dots,  \lambda_0\idMat_{d_l} + \lambda_{l_\times}\penMat_l), \label{eq:block-struct}
\end{align}
where $\design_{l,k}$ are the distributed design matrices of $b_l$ on sites $\iSite=1,\dots,\nSites$.

\subsection{Fitting Algorithm}\label{subsec:dist-fitting-algo}

We now describe the adaptions required to allow for distributed computations of the CWB fitting routine. In Sections~\ref{subsubsec:shared-effects} and~\ref{subsubsec:site-effects}, we show the equality between our distributed fitting approach and CWB fitted on pooled data. Section~\ref{subsec:distr-sse-blsel-pr} describes the remaining details such as distributed SSE calculations, distributed model updates, and pseudo residual updates in the distributed setup. Section~\ref{subsec:dist-cwb-algo} summarizes the distributed CWB algorithm and Section~\ref{subsec:dist-cwb-comm-cost} elaborates on the communication costs of our algorithm.

\subsubsection{Distributed Shared Effects Computation}\label{subsubsec:shared-effects}

Fitting CWB in a distributed fashion requires adapting the fitting process of the base learner $b_l$ in Algorithm~\ref{algo:cwb} to distributed data. To allow for shared effects computations across different sites without jeopardizing privacy, we take advantage of CWB's update scheme, which boils down to a (penalized) least squares estimation per iteration for every base learner. This allows us to build upon existing work such as \citet{karr2005secure} to fit linear models in a distributed fashion by just communicating aggregated statistics between sites and the host.  

In a first step, the aggregated matrices $\bm{F}_{l,\iSite} = \design_{l,\iSite}^\tran \design_{l,\iSite}$ and vectors $\bm{u}_{l,\iSite} = \design_{l,\iSite}^\tran \yv_\iSite$ are computed on each site. In our privacy setup (Section~\ref{subsec:dist-setup}), communicating $\bm{F}_{l,\iSite}$ and $\bm{u}_{l,\iSite}$ is allowed as long as the privacy-aggregation level per site is met. In a second step, the site information is aggregated to a global information $\bm{F}_l = \sum_{\iSite=1}^\nSites \bm{F}_{l,\iSite} + \penMat_l$ and $\bm{u}_l = \sum_{\iSite=1}^\nSites \bm{u}_{l,\iSite}$ and then used to estimate the model parameters $\thetabh_l = \bm{F}_l^{-1}\bm{u}_l$. This approach, referred to as $\operatorname{distFit}$, is explained again in detail in Algorithm~\ref{algo:lm-distr} and used for the shared effect computations of the model by substituting $\tbmh_l = \left(\design_l^\tran \design_l + \penMat_l\right)^{-1} \design^\tran_l \rmm$ (Algorithm~\ref{algo:cwb} line~\ref{algo:cwb:line:fitbl}) with $\tbmh_l = \operatorname{distFit}(\design_{l,1}, \dots, \design_{l,\nSites}, \rmm_1, \dots, \rmm_\nSites, \penMat_l)$. 

Note that the pseudo residuals $\rmm_k$ are also securely located at each site and are updated after each iteration. Details about the distributed pseudo residuals updates are explained in Section~\ref{subsec:distr-sse-blsel-pr}. We also note that the computational complexity of fitting CWB can be drastically reduced by pre-calculating and storing $(\design_{l}^\tran\design_{l} + \penMat_l)^{-1}$ in a first initialization step, as the matrix is independent of iteration $m$, and reusing these pre-calculated matrices in all subsequent iterations \citep[cf.~][]{schalk2022accelerated}. Using pre-calculated matrices also reduces the amount of required communication between sites and host.

\begin{algorithm}[H]
%\footnotesize
\small
\caption{Distributed Effect Estimation.\\ 
The line prefixes \doS and \doH indicate whether the operation is conducted at the sites (\doS) or at the host (\doH).}\label{algo:lm-distr} 
\vspace{0.15cm}
\hspace*{\algorithmicindent} \textbf{Input} Sites design matrices $\design_{l,1}, \dots, \design_{l,\nSites}$, \\
\hspace*{\algorithmicindent} \phantom{\textbf{Input} }response vectors $\yv_1, \dots, \yv_\nSites$ and\\
\hspace*{\algorithmicindent} \phantom{\textbf{Input} }an optional penalty matrix $\penMat_l$.\\
\hspace*{\algorithmicindent} \textbf{Output} Estimated parameter vector $\thetabh_l$.\vspace{0.15cm}
\hrule
\begin{algorithmic}[1]
\Procedure{$\operatorname{distFit}$}{$\design_{l,1}, \dots, \design_{l,\nSites}, \yv_1, \dots, \yv_\nSites, \penMat_l$}
    \For{$\iSite \in \{1, \dots, \nSites\}$}
        \State \doS $\bm{F}_{l,\iSite} = \design_{l,\iSite}^\tran \design_{l,\iSite}$
        \State \doS $\bm{u}_{l,\iSite} = \design_{l,\iSite}^\tran \yv_\iSite$
        \State \doS Communicate $\bm{F}_{l,\iSite}$ and $\bm{u}_{l,\iSite}$ to the host
    \EndFor
    \State \doH $\bm{F}_l = \sum_{\iSite=1}^\nSites \bm{F}_{l,\iSite} + \penMat_l$
    \State \doH $\bm{u}_l = \sum_{\iSite=1}^\nSites \bm{u}_{l,\iSite}$
    \State \doH \textbf{return} $\thetabh_l = \bm{F}_l^{-1}\bm{u}_l$
\EndProcedure
\end{algorithmic}
\end{algorithm}

\subsubsection{Distributed Site-specific Effects Computation}\label{subsubsec:site-effects}

If we pretend that the fitting of the base learner $b_{\lsite}$ is performed on the pooled data, we obtain
\begin{align}
    \thetabh_{\lsite}% &= \fitbl(\yv, \D, b_{\lsite}) 
    &= \left(\design_{\lsite}^\tran \design_{\lsite} +  \penMat_{\lsite}\right)^{-1}\design_{\lsite}^\tran \yv 
    %= \left(\begin{array}{c}
    %     \thetabh_{{\lsite}, 1}  \\
    %     \vdots \\
    %     \thetabh_{{\lsite}, \nSites} 
    %\end{array}\right) 
    \notag \\ 
    &= \left(
    \begin{array}{c}
         (\design_{l,1}^\tran \design_{l,1} + \lambda_0\idMat_{d_l} + \penMat_l)^{-1} \design_{l,1}^\tran \yv_1 \\
         \vdots \\
         (\design_{l,\nSites}^\tran \design_{l,\nSites} + \lambda_0\idMat_{d_l} + \penMat_l)^{-1} \design_{l,\nSites}^\tran \yv_\nSites
    \end{array}
    \right), \label{eq:bl-fit-kronecker-pooled}
\end{align}
where \eqref{eq:bl-fit-kronecker-pooled} is due to the block structure, as described in \eqref{eq:block-struct} of Section~\ref{subsec:distr-baselearner}. This shows that the fitting of the site-specific effects $\thetabh_{\lsite}$ can be split up into the fitting of individual parameters 
\begin{equation}\label{eq:shared-effect-per-site}
    \thetabh_{{\lsite},\iSite} = (\design_{l,\iSite}^\tran \design_{l,\iSite} + \lambda_0\idMat_{d_l} + \penMat_l)^{-1} \design_{l,\iSite}^\tran \yv_\iSite.
\end{equation}
It is thus possible to compute site-specific effects at the respective site without the need to share any information with the host. The host, in turn, only requires the SSE of the respective base learner (see next Section~\ref{subsec:distr-sse-blsel-pr}) to perform the next iteration of CWB. Hence, during the fitting process, the parameter estimates remain at their sites and are just updated if the site-specific base learner is selected. This again minimizes the amount of data communication between sites and host and speeds up the fitting process. After the fitting phase, the aggregated site-specific parameters are communicated once in a last communication step to obtain the final model.

\subsubsection{Pseudo Residual Updates, SSE Calculation, and Base Learner Selection}\label{subsec:distr-sse-blsel-pr}

The remaining challenges to run the distributed CWB algorithm are 1) the pseudo residual calculation (Algorithm~\ref{algo:cwb} line~\ref{algo:cwb:line:pr}), 2) the SSE calculation (Algorithm~\ref{algo:cwb} line~\ref{algo:cwb:line:sse}), and 3) base learner selection (Algorithm~\ref{algo:cwb} line~\ref{algo:cwb:line:blselection}). 

\paragraph{Distributed pseudo residual updates} The site-specific response vector $\yv_\iSite$ containing the values $y^{(i)},\ i \in \{1, \dots, n_\iSite\}$ is the basis of the pseudo residual calculation. We assume that every site $\iSite$ has access to all shared effects as well as the site-specific information of all site-specific base learners $b_{\lsite}$ only containing the respective parameters $\thetabh_{{\lsite},\iSite}$. Based on these base learners, it is thus possible to compute a \emph{site model} $\fmh_\iSite$ as a representative of $\fmh$ on every site $\iSite$. The pseudo residual updates $\rmm_\iSite$ per site are then based on $\fmh_\iSite$ via $\rmi_\iSite = -\nabla_f L(y^{(i)}, f(\xi))\vert_{f = \fmdh_\iSite},\  i \in \{1, \dots, n_\iSite\}$ using $\D_\iSite$. Most importantly, all remaining steps of the distributed CWB fitting procedure do not share the pseudo residuals $\rmm_\iSite$ in order to avoid information leakage about $\yv_\iSite$. 

\paragraph{Distributed SSE calculation and base learner selection} After fitting all base learners $b_l\in\mathcal{B}$ and $b_{\lsite}\in\mathcal{B}_\times$ to $\rmm_\iSite$, we obtain $\tbmh_l$, $l=1, \dots, \lvert\mathcal{B}\rvert$, and $\tbmh_{\lsite}$, $\lsite = 1_\kron, \dots, \lvert\mathcal{B}_\kron\rvert$. Calculating the SSE distributively for the $l\th$ and $\lsite\th$ base learner $b_l$ and $b_{\lsite}$, respectively, requires calculating $2\nSites$ site-specific SSE values:
\begin{align*}
    \sse_{l,\iSite} &= \sum_{i=1}^{n_\iSite} \left(\rmi_\iSite - b_l(\xi_\iSite, \tbmh_l)\right)^2 \\
    &= \sum_{i=1}^{n_\iSite} (\rmi - g_l(\xi)^\tran \tbmh_l)^2 \label{eq:distr-sse1}, \\
    \sse_{\lsite,\iSite} &= \sum_{i=1}^{n_\iSite} \left(\rmi_\iSite - b_{\lsite}(\xi_\iSite, \tbmh_{\lsite})\right)^2 \\
    &= \sum_{i=1}^{n_\iSite} (\rmi_\iSite - g_l(\xi)^\tran \tbmh_{\lsite, \iSite})^2. %\label{eq:distr-sse2}
\end{align*}
The site-specific SSE values are then sent to the host and aggregated to $\sse_l = \sum_{\iSite=1}^\nSites \sse_{l,\iSite}$. If privacy constraints have been met in all previous calculations, sharing the individual SSE values is not critical and does not violate any privacy constraints as the value is an aggregation of all $n_\iSite$ observations for all sites $\iSite$. 

Having gathered all SSE values at the host location, selecting the best base learner in the current iteration is done in the exact same manner as for the non-distributed CWB algorithm by selecting $l^{[m]} = \argmin_{l\in\{1, \dots, \lvert\mathcal{B}\rvert, 1_\times, \dots, \lvert\mathcal{B}\rvert_\times\}} \sse_l$. After the selection, the index $l^{[m]}$ is shared with all sites to enable the update of the site-specific models $\fmh_\iSite$. If a shared effect is selected, the parameter vector $\tbmh_{l^{[m]}}$ is shared with all sites. Caution must be taken when the number of parameters of one base learner is equal to the number of observations, as this allows reverse-engineering private data. In the case of a site-specific effect selection, no parameter needs to be communicated, as the respective estimates are already located at each site.

\begin{algorithm*}[!ht]
%\footnotesize
\small
\caption{Distributed CWB Algorithm.\\ 
The line prefixes \doS and \doH indicate whether the operation is conducted at the sites (\doS) or at the host (\doH).}\label{algo:cwb-distr}
\vspace{0.15cm}
\hspace*{\algorithmicindent} \textbf{Input} Sites with site data $\D_k$, learning rate $\nu$, number of boosting iterations $M$, loss\\
\hspace*{\algorithmicindent} \phantom{\textbf{Input} }function $L$, set of shared effects $\mathcal{B}$ and respective site-specific effects $\mathcal{B}_\times$\\
\hspace*{\algorithmicindent} \textbf{Output} Prediction model $\fh$\vspace{0.15cm}
\hrule

\begin{algorithmic}[1]
\Procedure{$\operatorname{distrCWB}$}{$\nu,L,\mathcal{B}, \mathcal{B}_\times$}
    \State Initialization: 
    \State \doH Initialize shared model $\fh^{[0]}_{\text{shared}}(\xv) = \argmin_{c\in\R}\riske(c)$
    \State \doS Calculate $\design_{l,\iSite}$ and $\bm{F}_{l,\iSite} = \design_{l,\iSite}^\tran \design_{l,\iSite}$, $\forall l\in\{1, \dots, \lvert\mathcal{B}\rvert\},\ \iSite\in\{1, \dots, \nSites\}$
    \State \doS Set $\fh^{[0]}_\iSite = \fh^{[0]}_{\text{shared}}$

    \For{$m \in \{1, \dots, M\}$ or \textbf{while} an early stopping criterion is not met}
        \State \doS Update pseudo residuals: 
        \State \doS\algospace$\rmi_\iSite = -\nabla_f L(y^{(i)}, f(\xi))\vert_{f = \fmdh_\iSite},\ \ \forall i \in \{1, \dots, n_\iSite\}$\label{algo:dist-cwb:line:pr}
        \For{$l \in \{1, \dots, \lvert\mathcal{B}\rvert\}$}
        
            \State \doH Calculate shared effect: $\tbmh_l = \operatorname{distFit}(\design_{l,1}, \dots, \design_{l,\nSites}, \yv_1, \dots, \yv_\nSites, \penMat_l)$
            \State \doH Communicate $\tbmh_l$ to the sites

            \For{$k \in \{1, \dots, \nSites\}$}
                \State \doS Fit $l\th$ site-specific effect: $\tbmh_{\lsite,\iSite} = (\bm{F}_{l,\iSite} + \lambda_0\idMat_{d_l} + \penMat_l)^{-1}\design_{l,\iSite}\rmm_\iSite$ 
                \State \doS Calculate the SSE for the $l\th$ shared and site-specific effect:
                \State \doS \algospace$\sse_{l,\iSite} = \sum_{i=1}^{n_\iSite} (\rmi - g_l(\xi)^\tran \tbmh_l)^2$                
                \State \doS \algospace$\sse_{\lsite,\iSite} = \sum_{i=1}^{n_\iSite} (\rmi_\iSite - g_l(\xi)^\tran \tbmh_{\lsite, \iSite})^2$
                \State \doS Send $\sse_{l,\iSite}$ and $\sse_{\lsite, \iSite}$ to the host
            \EndFor
            
            \State \doH Aggregate SSE values: $\sse_{l} = \sum_{\iSite=1}^\nSites \sse_{l,\iSite}$ and $\sse_{\lsite} = \sum_{\iSite=1}^\nSites \sse_{\lsite,\iSite}$
        \EndFor
        \State \doH Select best base learner: $l^{[m]} = \argmin_{l\in\{1, \dots, \lvert\mathcal{B}\rvert, 1_\times, \dots, \lvert\mathcal{B}\rvert\times\}} \sse_l$

        \If{$b_{l^{[m]}}$ is a shared effect}
            \State \doH Update model: $\fmh_{\text{shared}}(\xv) = \fmdh_{\text{shared}}(\xv) + \nu b_{l^{[m]}}(\xv, \tbmh_{l^{[m]}})$
            \State \doH Upload model update $\tbmh_{l^{[m]}}$ to the sites.
        \EndIf
        \State \doS Update site model $\fh_\iSite^{[m]}$ via parameter updates $\tbh_{l^{[m]}} = \tbh_{l^{[m]}} + \nu\tbmh_{l^{[m]}}$
    \EndFor
    \State \doS Communicate site-specific effects $\tbh_{1_\times}, \dots, \tbh_{\lvert\mathcal{B}\rvert_\times}$ to the host
    \State \doH Add site-specific effects to the model of shared effects $\fh^{[M]}_{\text{shared}}$ to obtain the full model $\fh^{[M]}$ 
    \State \doH\textbf{return} $\fh = \fh^{[M]}$
\EndProcedure
\end{algorithmic}
\end{algorithm*}

\subsection{Distributed CWB Algorithm with Site-Specific Effects}\label{subsec:dist-cwb-algo}

Assembling all pieces, our distributed CWB algorithm is summarized in Algorithm~\ref{algo:cwb-distr}.

\subsection{Communication Costs}\label{subsec:dist-cwb-comm-cost}

While the CWB iterations themselves can be performed in parallel on every site and do not slow down the process compared to a pooled calculation, it is worth discussing the communication costs of $\operatorname{distrCWB}$. During the initialization, data is shared just once, while the fitting phase requires the communication of data in each iteration. Let $d = \max_l d_l$ be the maximum number of basis functions (or, alternatively, assume $d$ basis functions for all base learners). The two main drivers of the communication costs are the number of boosting iterations $M$ and the number of base learners $\lvert\mathcal{B}\rvert$. Because of the iterative nature of CWB with a single loop over the boosting iterations, the communication costs (both for the host and each site) scale linearly with the number of boosting iterations $M$, i.e., $\mathcal{O}(M)$. For the analysis of communication costs in terms of the number of base learners, we distinguish between the initialization phase and the fitting phase.

\paragraph{Initialization} As only the sites share $\bm{F}_{l,\iSite}\in\R^{d\times d},\ \forall l\in\{1, \dots, \lvert\mathcal{B}\rvert\}$, the transmitted amount of values is $d^2\lvert\mathcal{B}\rvert$ for each site and therefore scales linearly with $\lvert\mathcal{B}\rvert$, i.e., $\mathcal{O}(\lvert\mathcal{B}\rvert)$. The host does not communicate any values during the initialization. 

\paragraph{Fitting} In each iteration, every site shares its vector $\design_{l,\iSite}^\tran \rmm_\iSite\in\R^d,\ \forall l\in\{1, \dots, \lvert\mathcal{B}\rvert\}$. Over the course of $M$ boosting iterations, each site therefore shares $dM\lvert\mathcal{B}\rvert$ values. Every site also communicates the SSE values, i.e., $2$ values (index and SSE value) for every base learner and thus $2M\lvert\mathcal{B}\rvert$ values for all iterations and base learners. In total, each site communicates $M\lvert\mathcal{B}\rvert(d + 2)$ values. The communication costs for all sites are therefore $\mathcal{O}(\lvert\mathcal{B}\rvert)$. The host, in turn, communicates the estimated parameters $\tbmh\in\R^d$ of the $\lvert\mathcal{B}\rvert$ shared effects. Hence, $dM\lvert\mathcal{B}\rvert$ values as well as the index of the best base learner in each iteration are transmitted. In total, the host therefore communicates $dM\lvert\mathcal{B}\rvert + M$ values to the sites, and costs are therefore also $\mathcal{O}(\lvert\mathcal{B}\rvert)$.

\section{Application}\label{sec:application}

We now showcase our algorithm on a heart disease data set that consists of patient data gathered all over the world. The data were collected at four different sites by the 1) Hungarian Institute of Cardiology, Budapest (Andras Janosi, M.D.), 2) University Hospital, Zurich, Switzerland (William Steinbrunn, M.D.), 3) University Hospital, Basel, Switzerland (Matthias Pfisterer, M.D.), and 4) V.A. Medical Center, Long Beach, and Cleveland Clinic Foundation (Robert Detrano, M.D., Ph.D.), and is thus suited for a multi-site distributed analysis. The individual data sets are freely available at \url{https://archive.ics.uci.edu/ml/datasets/heart+disease}~\citep{uciMLrepository}. For our analysis, we set the privacy level (cf.~Section~\ref{subsec:dist-setup}) to 5 which is a common default. 

\subsection{Data Description}

The raw data set contains 14 covariates, such as the chest pain type (\texttt{cp}), resting blood pressure (\texttt{trestbps}), maximum heart rate (\texttt{thalach}), sex, exercise-induced angina (\texttt{exang}), or ST depression (i.e., abnormal difference of the ST segment from the baseline on an electrocardiogram) induced by exercise relative to rest (\texttt{oldpeak}). A full list of covariates and their abbreviations is given on the data set's website. After removing non-informative covariates and columns with too many missing values at each site, we obtain $n_{\text{cleveland}} = 303$, $n_{\text{hungarian}} = 292$, $n_{\text{switzerland}} = 116$, and $n_{\text{va}} = 140$ observations and 8 covariates. A table containing the description of the abbreviations of these covariates is given in Table~1 in the Supplementary Material~B.1. For our application, we assume that missing values are completely at random and all data sets are exclusively located at each sites. The task is to determine important risk factors for heart diseases. The target variable is therefore a binary outcome indicating the presence of heart disease or not. 

\subsection{Analysis and Results}\label{subsec:analysis-res}

\newcommand{\mstop}{5578\xspace}
\newcommand{\nselshared}{782\xspace}
\newcommand{\nselsite}{4796\xspace}
\newcommand{\patience}{5\xspace}
\newcommand{\dfMain}{2.2\xspace}
\newcommand{\dfRI}{3\xspace}
\newcommand{\learningrate}{0.1\xspace}
\newcommand{\nknots}{10\xspace}

We follow the practices to setup CWB as mentioned in Section~\ref{subsubsec:fitting-algo} and run the distributed CWB algorithm with a learning rate of \learningrate and a maximum number of 100000 iterations. To determine an optimal stopping iteration for CWB, we use 20 \% of the data as validation data and set the patience to \patience iterations. In other words, the algorithm stops if no risk improvement on the validation data is observed in \patience consecutive iterations. For the numerical covariates, we use a P-spline with \nknots cubic basis functions and second-order difference penalties. All base learners are penalized accordingly to a global degree of freedom that we set to \dfMain (to obtain unbiased feature selection) while the random intercept is penalized according to \dfRI degrees of freedom (see the Supplementary Material~B.2 for more details). Since we are modelling a binary response variable, $h^{-1}$ is the inverse logit function $\operatorname{logit}^{-1}(f) = (1 + \exp(-f))^{-1}$. The model for an observation of site $s$, conditional on its random effects $\bm{\gamma}$, is given in the Supplementary Material~B.3.
% \begin{align*}
% h(\mathbb{E}(&Y\mid\bm{x},\bm{u}, \bm{\gamma}_s)) = \mathds{1}(x_{\text{sex}}=\text{``male''}) \beta_{\text{sex}} + \mathds{1}(x_{\text{exang}}=\text{``yes''}) \beta_{\text{exang}} + \\
% &\quad \textstyle \sum_{i\in\{1,2,3\}}\mathds{1}(x_{\text{cp}}=i) \beta_{\text{cp},i} + \textstyle  \sum_{i\in\{1,2\}}\mathds{1}(x_{\text{restecg}}=i) \beta_{\text{restecg},i} + \\
% &\quad \mathds{1}(x_{\text{sex}}=\text{``male''}) \gamma_{\text{sex},s} + \textstyle \sum_{i\in\{1,2,3\}} \mathds{1}(x_{\text{cp}}=i) \gamma_{\text{cp},i,s}  + \\
% &\quad \textstyle \sum_{i\in\{1,2\}} \mathds{1}(x_{\text{restecg}}=i) \gamma_{\text{resecg},i,s}  + \mathds{1}(x_{\text{exang}}=\text{``yes''}) \gamma_{\text{exang},s} + \gamma_{0,s} + \\
% &\quad\phi_{\text{age}}(x_{\text{age}}) + \phi_{\text{trestbps}}(x_{\text{trestbps}})+ \phi_{\text{thalach}}(x_{\text{thalach}}) + \phi_{\text{oldpeak}}(x_{\text{oldpeak}}) +\\
% &\quad\phi_{\text{age},s}(x_{\text{age}}) + \phi_{\text{trestbps},s}(x_{\text{trestbps}})+ \phi_{\text{thalach},s}(x_{\text{thalach}}) + \phi_{\text{oldpeak},s}(x_{\text{oldpeak}}).
% \end{align*}

\begin{figure}[H]
\centering
\includegraphics[width=\columnwidth]{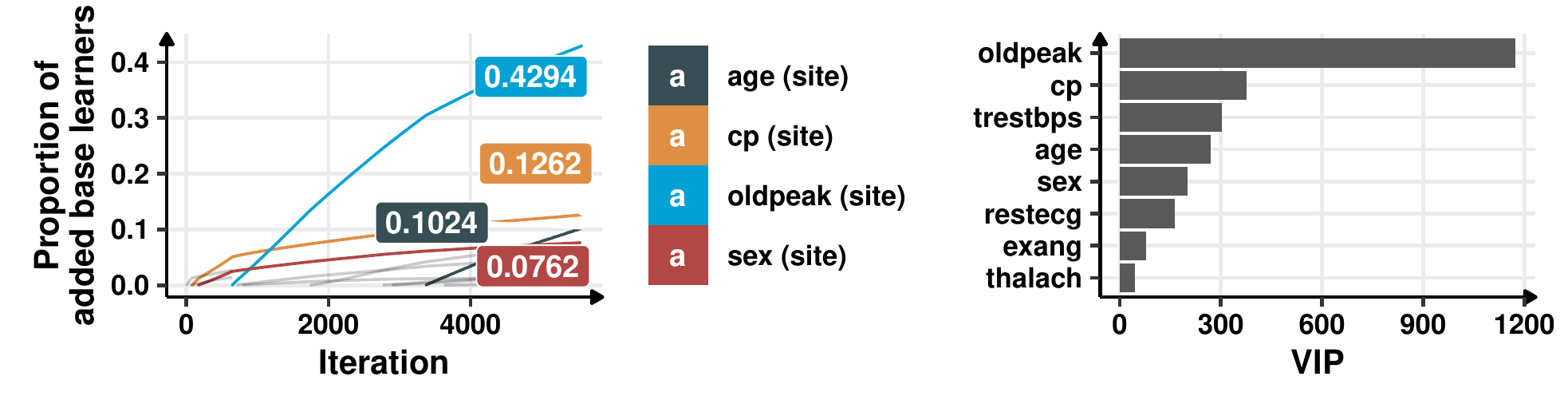}
%\caption{Left: Model trace showing how and when additive terms entered the model. CWB starts by including the site effect of \texttt{exang} and \texttt{cp}. At later stages, the site effect of \texttt{oldpeak} is included as well as the site effects of \texttt{restecg} as well as the shared effect of \texttt{trestbps}. Right: Variable importance \citep[cf.][]{au2019component} of selected features in decreasing order. The contribution to the overall empirical risk reduction is strongest for \texttt{oldpeak}, followed by \texttt{cp}, \texttt{trestbps}, and \texttt{age}.}
\caption{Left: Model trace showing how and when the four most selected additive terms entered the model. Right: Variable importance \citep[cf.][]{au2019component} of selected features in decreasing order.}
\label{fig:bl-vip}
\end{figure}

\paragraph{Results} The algorithm stops after $m_{\text{stop}} = \mstop$ iterations as the risk on the validation data set starts to increase (cf. Figure~1 in the Supplementary Material~B.4). Out of these \mstop iterations, the distributed CWB algorithm selects a shared effect in \nselshared iterations and site-specific effects in \nselsite iterations. This indicates that the data is rather heterogeneous and requires site-specific (random) effects. Figure~\ref{fig:bl-vip} (Left) shows traces of how and when the different additive terms (base learners) entered the model during the fitting process and illustrates the selection process of CWB. 

\begin{figure}[h]
    \centering
    \includegraphics[trim={0 0.2cm 0 0},width=\columnwidth]{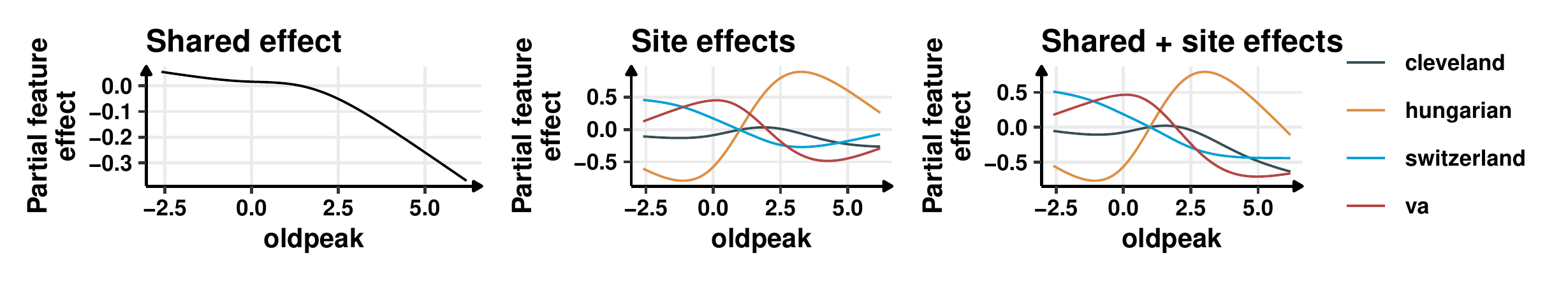}
    \caption{Decomposition of the effect of \texttt{oldpeak} into the shared (left) and the site-specific effects (middle). The plot on the right-hand side shows the sum of shared and site-specific effects.}
    \label{fig:fe-oldpeak}
\end{figure}

The estimated effect of the most important feature \texttt{oldpeak} (cf. Figure~\ref{fig:bl-vip}, Right) found is further visualized in Figure~\ref{fig:fe-oldpeak}. Looking at the shared effect, we find a negative influence on the risk of heart disease when increasing ST depression (\texttt{oldpeak}). When accounting for site-specific deviations, the effect becomes more diverse, particularly for Hungary.
 
In the Supplementary Material~B.5 and~B.6, we provide the partial effects for all features and showcase the conditional predictions of the fitted GAMM model for a given site. 

\paragraph{Comparison of Estimation Approaches} The previous example shows partial feature effects that exhibit shrinkage due to the early stopping of CWB's fitting routine. While this prevents overfitting and induces a sparse model, we can also run CWB for a very large amount of iterations without early stopping to approximate the unregularized and hence unbiased maximum likelihood solution. We illustrate this in the following by training CWB and our distributed version for 100000 iterations and compare its partial effects to the ones of a classical mixed model-based estimation routine implemented in the \texttt{R} package \texttt{mgcv}~\citep{Wood.2017.book}.

\begin{figure}[!h]
    \centering
    \includegraphics[trim={0 0.3cm 0 0}, width=\columnwidth]{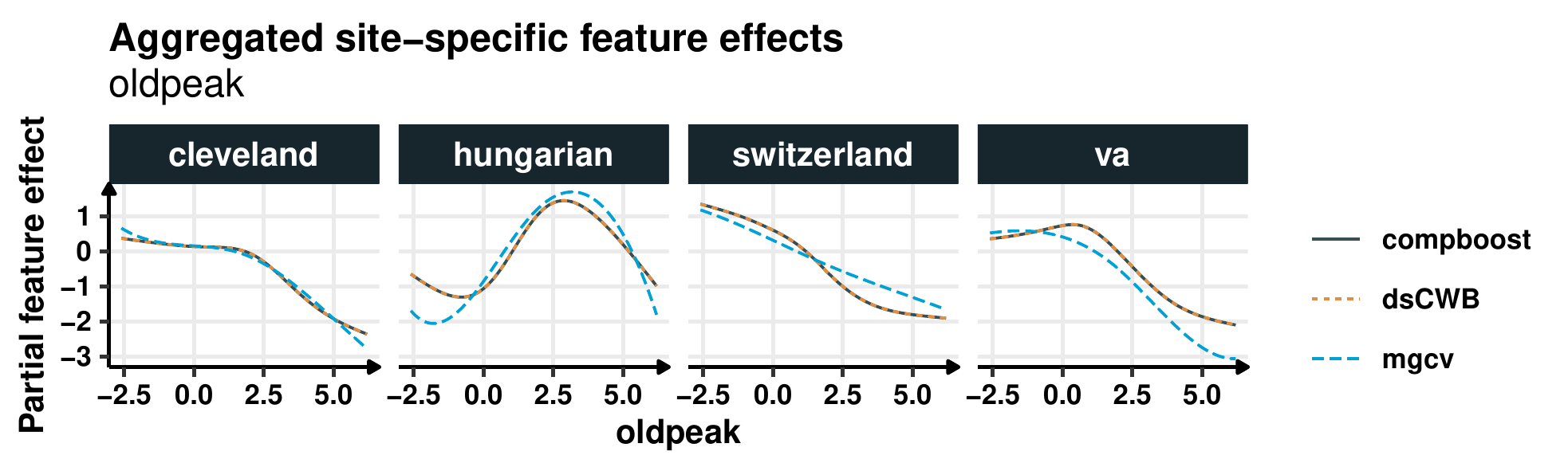}
    \caption{Comparison of the site-specific effects for \texttt{oldpeak} between the distributed (\texttt{dsCWB}) and pooled CWB approach (\texttt{compboost}) as well as estimates of from \texttt{mgcv}.}
    \label{fig:comparison-mgcv}
\end{figure}

Results of the estimated partial effects of our distributed CWB algorithm and the original CWB on pooled data show a perfect overlap (cf. Figure~\ref{fig:comparison-mgcv}). This again underpins the lossless property of the proposed algorithm. The site-specific effects on the pooled data are fitted by defining a row-wise Kronecker base learner for all features and the site as a categorical variable. The same approach is used to estimate a GAMM using \texttt{mgcv} fitted on the pooled data with tensor products between the main feature and the categorical site variable. A comparison of all partial feature effects is given in the Supplementary Material~B.7 showing good alignment between the different methods. For the \texttt{oldpeak} effect shown in Figure~\ref{fig:comparison-mgcv}, we also see that the partial effects of the two CWB methods are very close to the mixed model-based estimation, with only smaller differences caused by a slightly different penalization strength of both approaches. The empirical risk is $0.4245$ for our distributed CWB algorithm, $0.4245$ for CWB on the pooled data, and $0.4441$ for the GAMM on the pooled data.

\section{Discussion}

We proposed a novel algorithm for distributed, lossless, and privacy-preserving GAMM estimation to analyze horizontally partitioned data. To account for data heterogeneity of different sites we introduced site-specific (smooth) random effects. Using CWB as the fitting engine allows estimation in high-dimensional settings and fosters variable as well as effect selection. This also includes a data-driven selection of shared and site-specific features, providing additional data insights. Owing to the flexibility of boosting and its base learners, our algorithm is easy to extend and can also account for interactions, functional regression settings~\citep{brockhaus2017boosting}, or modeling survival tasks~\citep{bender2020general}.

%% OLD
An open challenge for the practical use of our approach is its high communication costs. For larger iterations (in the 10 or 100 thousands), computing a distributed model can take several hours. One option to reduce the total runtime is to incorporate accelerated optimization recently proposed in \citet{schalk2022accelerated}. Another driver that influences the runtime is the latency of the technical setup. Future improvements could reduce the number of communications, e.g., via multiple fitting rounds at the different sites before communicating the intermediate results. 

%An open challenge for the practical use of our approach is its high communication costs. For larger iterations (in the 10 or 100 thousands), computing a distributed model can take several hours. 
%A driver that influences the runtime is the latency of the technical setup. Future improvements could reduce the number of communications, e.g., via multiple fitting rounds at the different sites before communicating the intermediate results.
%Despite being a quasi-tuning-free approach, it can be beneficial in terms of the model's performance to automatically select HPs via tuning which also benefits from a faster fitting.
%To this end, an example of how to apply tuning to CWB and extensions to accelerate CWB's model optimization that can be directly transferred to the distributed CWB algorithm can be viewed in \cite{schalk2022accelerated}.

A possible future extension of our approach is to account for both horizontally and vertically distributed data. Since the algorithm is performing component-wise (coordinate-wise) updates, the extension to vertically distributed data naturally falls into the scope of its fitting procedure. This would, however, require a further advanced technical setup and the need to ensure consistency across sites.

\backmatter

\section*{Declarations}

The authors declare that they have no known competing financial interests or personal relationships that could have appeared to influence the work reported in this paper.

\bibliography{references}

\end{document}